\documentclass[10pt,twocolumn,letterpaper]{article}

\usepackage{cvpr}              

\usepackage{graphicx}
\usepackage{amsmath}
\usepackage{amssymb}
\usepackage{multirow}
\usepackage{booktabs}
\usepackage{url}

\usepackage[pagebackref,breaklinks,colorlinks]{hyperref}

\usepackage[capitalize]{cleveref}
\crefname{section}{Sec.}{Secs.}
\Crefname{section}{Section}{Sections}
\Crefname{table}{Table}{Tables}
\crefname{table}{Tab.}{Tabs.}

\def\confName{CVPR}
\def\confYear{2022}

\begin{document}

\title{Learnable Locality-Sensitive Hashing for Video Anomaly Detection}
\author{
Yue Lu \thanks{Equal contribution.} \and Congqi Cao \footnotemark[1]~\textsuperscript{~,}\thanks{Corresponding author.}\and Yanning Zhang \\
\and School of Computer Science, Northwestern Polytechnical University\\
{\tt\small {zugexiaodui@mail.nwpu.edu.cn, \{congqi.cao, ynzhang\}@nwpu.edu.cn}}
}

\maketitle

\begin{abstract}
Video anomaly detection (VAD) mainly refers to identifying anomalous events that have not occurred in the training set where only normal samples are available.
Existing works usually formulate VAD as a reconstruction or prediction problem. However, the adaptability and scalability of these methods are limited.
In this paper, we propose a novel distance-based VAD method to take advantage of all the available normal data efficiently and flexibly.
In our method, the smaller the distance between a testing sample and normal samples, the higher the probability that the testing sample is normal.
Specifically, we propose to use locality-sensitive hashing (LSH) to map samples whose similarity exceeds a certain threshold into the same bucket in advance.
In this manner, the complexity of near neighbor search is cut down significantly.
To make the samples that are semantically similar get closer and samples not similar get further apart, we propose a novel learnable version of LSH that embeds LSH into a neural network and optimizes the hash functions with contrastive learning strategy. The proposed method is robust to data imbalance and can handle the large intra-class variations in normal data flexibly. Besides, it has a good ability of scalability. Extensive experiments demonstrate the superiority of our method, which achieves new state-of-the-art results on VAD benchmarks.
\end{abstract}

\section{Introduction}

Video anomaly detection (VAD) aims to identify anomalous events that do not conform to the expectation, where typically only normal samples are available during training \cite{SurveySingleScene2020ramachandra}. It plays an important role in intelligent surveillance.
However, it is a highly challenging task for the following reasons.
First, anomalous events are defined according to circumstances, thus the same activity may be normal or anomalous in different scenes.
Second, traditional supervised binary-classification methods are inapplicable to VAD, due to the absence of anomalous events in training set \cite{AbnormalEvent2013lu, RevisitSparse2017luo, MultitimescaleTrajectory2020rodrigues}.
Third, some kinds of normal events happen frequently while some happen occasionally. Therefore, diverse training data has large intra-class variations and imbalanced distribution.

Prevalent VAD methods can be classified into distance-based, reconstruction-based and prediction-based methods.
Distance-based methods detect anomalies according to the distance between testing data and representations of all the normal data. Representative methods are based on video patches \cite{StreetScene2020ramachandra, LearningDistance2020ramachandra}, decision boundaries of one-class SVM \cite{ObjectCentricAutoEncoders2019ionescu, DetectingAbnormal2019ionescu} and cluster centers \cite{ClusteringDriven2020chang, ClusterAttention2020wang, GraphEmbedded2020markovitz}.
However, these existing methods are not adaptive enough since some essential settings (\eg the number of clusters) are determined artificially. Besides, they suffer from large computation cost.
Reconstruction-based methods \cite{LearningTemporal2016hasan, LatentSpace2019abati, AnomalyDetection2019nguyen, VideoAnomaly2020fan, MemorizingNormality2019gong} and prediction-based methods \cite{FutureFrame2018liu, LearningRegularity2019morais, MultitimescaleTrajectory2020rodrigues, AppearanceMotionMemory2021cai, AnomalyDetection2021georgescu, PoseCVAEAnomalous2021jain, LearningMemoryGuided2020park, LearningNormal2021lv, HybridVideo2021liu} train auto-encoders on normal data to reconstruct the current frame or to predict the next frame. They assume that anomalous frames are hard to reconstruct or predict, hence anomalies can be detected according to high reconstruction or prediction errors.
The main drawback of these methods is that they do not consider the diversity of normal patterns explicitly \cite{LearningMemoryGuided2020park}. Hence, the adaptability and scalability of these methods to complex scenarios are limited.
To tackle this problem, a number of recent works augment reconstruction-based or prediction-based methods with distance-based methods, such as \cite{MemorizingNormality2019gong}, \cite{LearningMemoryGuided2020park}, \cite{LearningNormal2021lv}, \cite{AppearanceMotionMemory2021cai} and \cite{HybridVideo2021liu}. Concretely, they design a memory module to store prototypical normal patterns in training phase and retrieve them in testing phase for reconstruction or prediction.

\begin{figure*}[tph]
	\centering
	\includegraphics[]{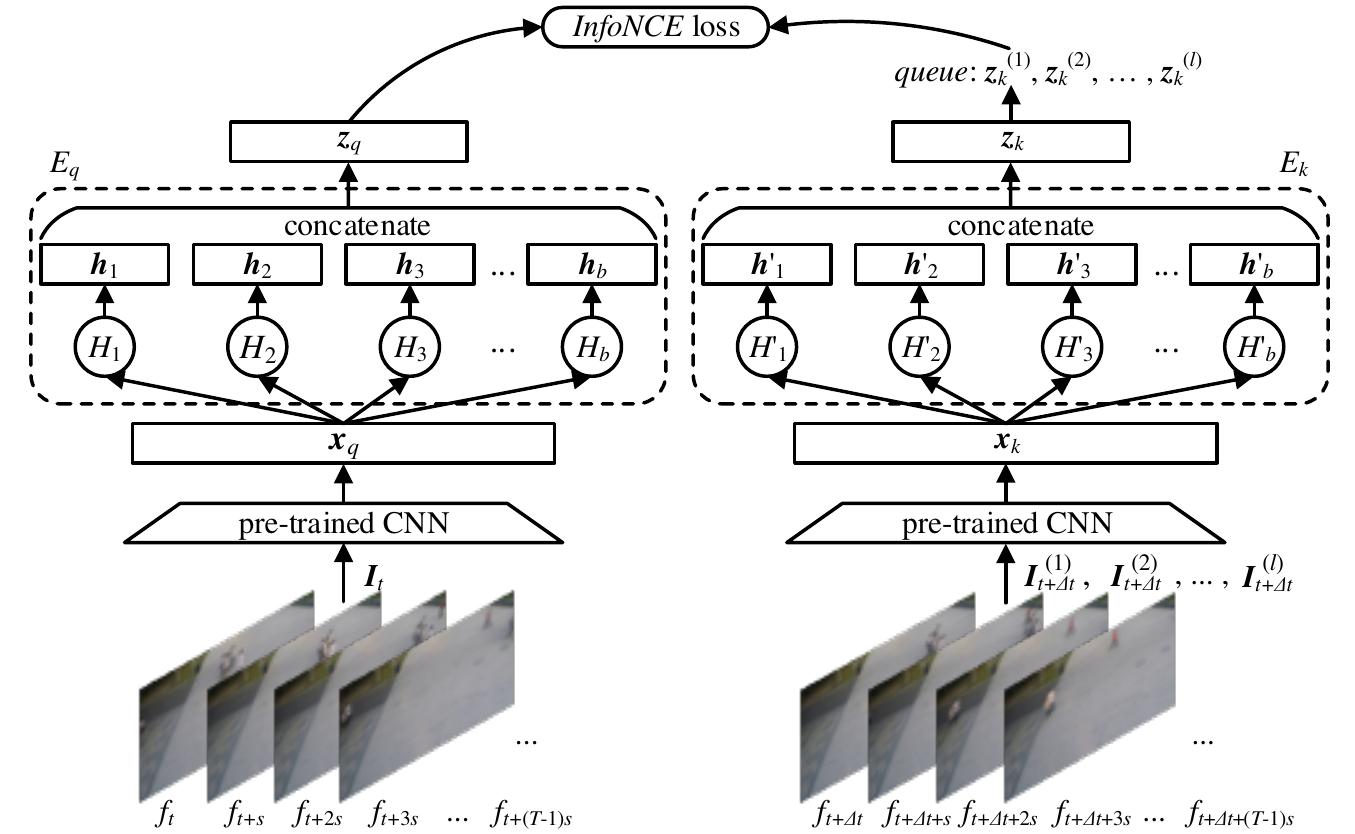}
	\caption{Overview of the proposed learnable locality-sensitive hashing. The architecture is divided into two symmetric parts. For each part, a pre-trained CNN extracts a feature $\boldsymbol{x}$ of the sampled video snippet $\boldsymbol{I}$ at first. Then, $\boldsymbol{x}$ is fed into an encoder $E$ and generates a low-dimensional hash code $\boldsymbol{z}$. The encoder is composed of $b$ parallel hash layers. A total of $l$ $\boldsymbol{z}_k$s are put in a queue to participate in the calculation of \textit{InfoNCE} loss with $\boldsymbol{z}_q$.
	}
	\label{fig1}
\end{figure*}

Considering the challenges of VAD, there is no sufficient and valid supervision knowledge in training phase to direct us what information we could discard. We intend to make full use of all the normal data for video anomaly detection.
Therefore, in this paper, we propose an efficient distance-based method, named Learnable Locality-Sensitive Hashing (LLSH), to distinguish anomalies from normal events.
Instead of directly comparing the distance between the testing sample and every training sample which would bring huge computational cost, we propose to use locality-sensitive hashing (LSH) to hash normal samples into buckets in advance, where samples whose similarity exceeds an approximate threshold are mapped into the same bucket.
During anomaly detection, the testing sample is also mapped into a bucket.
We only need to compute the distance between the testing sample and normal samples in the same bucket. As a result, the complexity of near neighbor search is cut down significantly.
Given that the representations of data are normally high-dimensional, we propose to compute the distances between low-dimensional hash codes instead of the original representations, which further reduces the computation cost.
Vanilla LSH uses random vectors in hash functions, which limits its adaptability for target datasets. 
To make the samples that are semantically similar get closer and samples not similar get further apart according to the known data structure, we propose a novel learnable locality-sensitive hashing (LLSH). It implements LSH as a parametric network layer and embeds it into an end-to-end trainable neural network.
Due to lack of explicit supervision information, we propose to optimize LLSH using contrastive learning strategy, where MoCo \cite{MomentumContrast2020he} is adopted in our framework.
Our proposed LLSH is shown in \cref{fig1}.
The architecture contains two symmetric parts. Each part is composed of a pre-trained CNN for feature extraction and a hash encoder. There is a set of parallel hash layers in the hash encoder to map the features to low-dimensional hash codes.
For contrastive learning, video snippets are fed into the symmetric network simultaneously.
A queue is maintained in the right part to calculate \textit{InfoNCE} loss among the hash codes stored in it and the current hash code output by the left part.
The parameters in the left part are updated through back-propagation, while those in the right part are updated by momentum.
After the optimization, the hash functions in the hash layers are tuned to be more adaptive to target datasets, hence improving the performance for VAD.

Our proposed LLSH can be regarded as an advanced version of KNN, which adaptively sets the number of nearest neighbors according to similarity thresholds.
In LLSH, a bucket is actually a cluster of similar samples. Therefore, it also can be seen as an advanced version of K-means, where the number of clusters (\ie buckets) are adaptively determined by the distribution of normal data.
In addition to the superior of adaptability, LLSH is much more efficient compared with KNN and K-means. Its computation cost is only $0.26\%$ of KNN and $3.00\%$ of K-means in our experiments.
Compared with memory modules, LLSH retains the whole knowledge of training data.
Hence, it is robust to data imbalance and can handle the large intra-class variations in normal data flexibility, which outperforms recent memory-augmented prediction methods by $4\% \sim 7\%$ in experiments.
Besides, LLSH has a good ability of scalability for newly added data, leading competing methods by $2\% \sim 6\%$.

We conduct extensive experiments on three VAD benchmarks, \ie, Avenue \cite{AbnormalEvent2013lu}, ShanghaiTech \cite{RevisitSparse2017luo} and Corridor \cite{MultitimescaleTrajectory2020rodrigues} datasets.
Our proposed LLSH achieves new state-of-the-art results on the three datasets in both micro-AUC and macro-AUC metrics.
Especially for the datasets with complex scenarios, our method surpasses existing methods with a large margin.
The efficiency, adaptability and scalability of LLSH are fully verified.
Our code is available at \url{https://github.com/zugexiaodui/LLSHforVAD}.

In summary, our main contributions include:

\begin{itemize}
	\item We introduce locality-sensitive hashing (LSH) and develop its paradigm of calculating distance for video anomaly detection.
	\item We propose a learnable locality-sensitive hashing (LLSH) that embeds LSH in an end-to-end trainable neural network and tunes the hash functions automatically in a contrastive learning framework.
	\item Our proposed LLSH is superior in efficiency, adaptability and scalability, achieving new state-of-the-art performance on VAD datasets.
\end{itemize}

\section{Related Work}

\textbf{Distance-based Video Anomaly Detection.} 
Distance-based methods create a model to learn the representations of normal data, and measure deviations from the model to determine anomaly scores.
Video patches \cite{StreetScene2020ramachandra, LearningDistance2020ramachandra}, decision boundaries of one-class SVM \cite{LearningDeep2015xu, AnomalyDetection2017tran, DeepAppearance2017smeureanu, ObjectCentricAutoEncoders2019ionescu, DetectingAbnormal2019ionescu} and cluster centers \cite{ClusteringDriven2020chang, ClusterAttention2020wang, GraphEmbedded2020markovitz} are commonly used in distance-based methods.
For example, Ramachandra \etal \cite{LearningDistance2020ramachandra} build a concise representative exemplar set of normal video patches to reduce the number of training samples. They employ a simple nearest neighbor lookup for anomaly detection.
Wang \etal \cite{ClusterAttention2020wang} train a deep $k$-cluster model for normal data,
and detect the anomaly by the similarities between its representations and corresponding $k$ centers.
Compared with the above methods, our proposed LLSH takes full advantage of the knowledge of normal data. It has a stronger ability of adaptability to flexibly handle the large intra-class variations in normal data.

\textbf{Deep Hashing.}
Deep hashing is a combination of deep learning and hashing algorithms.
In the field of computer vision, it is widely used in image retrieval, such as \cite{DeepLearning2015lin}, \cite{DeepSupervised2016liu} and \cite{DeepLearning2016nguyen}.
They aim to produce similar binary codes for semantically similar data, which is the same as the motivation of our learning process.
However, VAD is more challenging since there is no supervision information.
We tackle this problem by contrastive learning strategy.

\textbf{Contrastive Learning.}
In self-supervised learning, contrastive learning approaches (\eg SimCLR \cite{SimpleFramework2020chen}, MoCo \cite{MomentumContrast2020he}) present promising performance for image representations.
They use contrastive loss functions to narrow the distance between a positive pair of transformed inputs from the same sample, and broaden the distances of negative pairs from different samples.
Feichtenhofer \etal \cite{LargeScaleStudy2021feichtenhofer} propose that temporally-persistent features in the same video are effective for video representations, which inspires the design of LLSH.
In VAD, Wang \etal \cite{ClusterAttention2020wang} adopt MoCo to drive the deep clustering. They use several spatial transformations for the same snippet to generate positive pairs, but ignore the temporal relations of different snippets. We use temporally nearby snippets as positive pairs, which is simpler and more efficient in practice.

\begin{figure*}[t]
	\centering
	\includegraphics{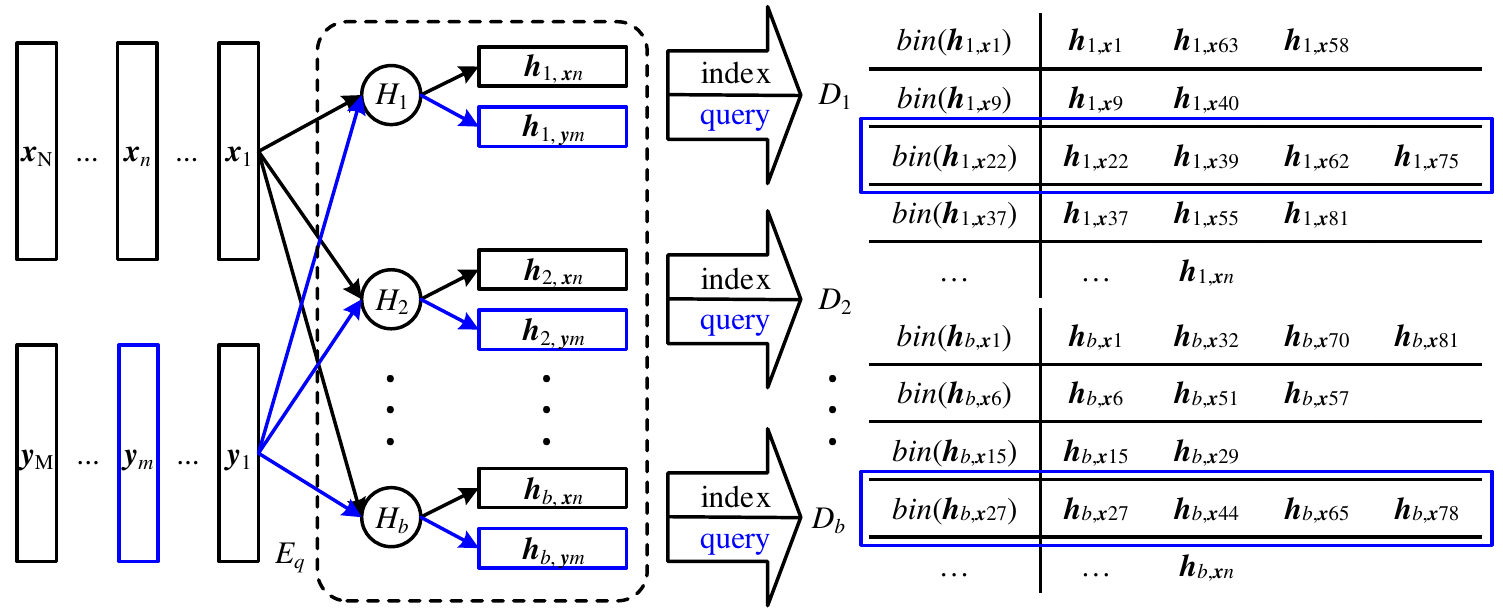}
	\caption{Overview of our anomaly detection, which includes two sequential stages: index and query. In the index stage, the hash codes of training features $\boldsymbol{x}_1, \cdots, \boldsymbol{x}_N$ are stored in $b$ parallel hash tables $D_1, \cdots, D_b$. In the query stage (in blue color), every testing feature $\boldsymbol{y}_i$ ($i \in \{1, \cdots M\}$) finds $b$ sets of similar hash codes.
	}
	\label{fig2}
\end{figure*}

\section{Method}

\subsection{Learnable Hashing}

In the training phase, we aim to train a hashing network that encodes video snippets to short hash codes. An overview of our learnable locality-sensitive hashing (LLSH) is shown in \cref{fig1}.
The left and right parts of the model have symmetric structures.

For the left part, firstly, a total of $T$ frames starting from the $t$-th frame are sampled as a video snippet $\boldsymbol{I}_t$ with a sampling-rate $s$.
Next, the snippet $\boldsymbol{I}_t$ is fed into a pre-trained CNN to extract a feature vector $\boldsymbol{x}_q$.
Then, we take $\boldsymbol{x}_q$ as the input of the hash encoder $E_q$, which contains $b$ parallel hash layers $H_1, H_2, \cdots H_b$.
In our implementation, they are fully-connected layers with $sigmoid$ function. Note that these fully-connected layers do not have bias terms.
Each layer $H_i$, $i\in \{1,2,\cdots,b\}$, maps $\boldsymbol{x}_q$ to a short hash code $\boldsymbol{h}_i$ of length $r$. Every bit of $\boldsymbol{h}_i$ gets a value in the range of $(0,1)$ because of the $sigmoid$ function.
All the $b$ short hash codes are concatenated as a long code $\boldsymbol{z}_q$ to participate in the calculation of loss.

The right part is symmetrical to the left and uses the same initialization.
A snippet $\boldsymbol{I}_{t+\varDelta t}$, in which $\varDelta t$ is a temporal offset, is sampled as the input for the right part.
The recent $l$ codes $\boldsymbol{z}_k^{(1)}, \boldsymbol{z}_k^{(2)}, \cdots, \boldsymbol{z}_k^{(l)}$ output by $E_k$ are put in a first-in-first-out (FIFO) queue. Finally, we compute the \textit{InfoNCE} loss \cite{RepresentationLearning2018oord} between $\boldsymbol{z}_q$ and the hash codes in the queue:
\begin{equation}
	L = - \log \frac{\exp(\boldsymbol{z}_q \cdot \boldsymbol{z}_{k+}/\tau)}{\sum_{i=0}^{l} \exp(\boldsymbol{z}_q \cdot \boldsymbol{z}_{k}^{(i)}/\tau)} ,
\end{equation}
where $\boldsymbol{z}_{k+}$ is the one output at the same time with $\boldsymbol{z}_q$, and $\tau$ is a temperature hyper-parameter, 

In the back-propagation phase, only the parameters $\theta_q$ of $E_q$ are updated. The parameters $\theta_k$ of $E_k$ is updated by momentum \cite{MomentumContrast2020he}:
\begin{equation}
	 \theta_k \leftarrow m \theta_k + (1-m) \theta_q ,
\end{equation}
where $m \in [0, 1)$ is a momentum coefficient. The parameters of the pre-trained CNN are freezed.

Via the above operations, the hashing process is transformed into a parametric and differentiable operation that can be embedded into neural networks and trained end-to-end with contrastive learning strategy.
We can get hash functions that are more adaptive to data distribution and perform anomaly detection based on it.

\subsection{Anomaly Detection}

In the anomaly detection phase, we aim to calculate the distances between the representations of a testing sample and its similar training samples. To this end, we propose to use the optimized locality-sensitive hashing to search similar items.

The process of our anomaly detection includes an index stage and a query stage. Only the left part of the model after training is used in both stages. In the index stage, all hash codes of training data are stored in hash tables. In the query stage, for each hash code of testing data, similar hash codes of training data are looked up from the hash tables. The anomaly score is determined by their distances. An illustration of the anomaly detection process is shown in \cref{fig2}.

\subsubsection{Index}
The features of training data extracted by the pre-trained CNN are denoted as $\boldsymbol{x}_1, \cdots, \boldsymbol{x}_N$, where $N$ represents the number of training samples.
First, The $n$-th feature $\boldsymbol{x}_n$ ($n \in \{1, \cdots, N\}$) is mapped to $b$ hash codes of $r$ bits $\boldsymbol{h}_{1,\boldsymbol{x}_n}, \cdots, \boldsymbol{h}_{b,\boldsymbol{x}_n}$ by the parallel hash layers $H_1, \cdots, H_b$ in $E_q$.
Next, we construct $b$ hash tables to store the representations (\ie hash codes) of $\boldsymbol{x}_1, \cdots, \boldsymbol{x}_N$.

We use $bin(\cdot) \in \{0,1\}^r$ to denote the binary function:
\begin{equation}
	bin(\boldsymbol{h}) = \begin{cases}
		1, \quad if ~ \boldsymbol{h}^{<i>} >= 0.5, \\
		0, \quad if ~ \boldsymbol{h}^{<i>} <0.5, 
	\end{cases}
	\mathrm{for~all~}i \in \{1, \cdots, r\}.
\end{equation}

In the $j$-th hash table $D_j$ ($j \in \{1, \cdots, b\}$) , for all $n \in \{1, \cdots, N\}$, $\boldsymbol{h}_{j,\boldsymbol{x}_n}$ and its binary code $bin(\boldsymbol{h}_{j,\boldsymbol{x}_n})$ are respectively stored as a value and a key. The values sharing the same key are stored in the same bucket.
As a result, each hash table stores $N$ hash codes with a number of binary-valued keys.

\subsubsection{Query}
The features of testing data are denoted as $\boldsymbol{y}_1, \cdots, \boldsymbol{y}_M$, where $M$ represents the number of testing samples.
We take the $m$-th ($m \in \{1, \cdots, M\}$) feature $\boldsymbol{y}_m$ for example to demonstrate the process and principle of our query stage.
First, it is also mapped to $b$ hash codes $\boldsymbol{h}_{1,\boldsymbol{y}_m}, \cdots, \boldsymbol{h}_{b,\boldsymbol{y}_m}$ by the hash layers.
We use the binary code $bin(\boldsymbol{h}_{j,\boldsymbol{y}_m})$ as the key to look up a bucket in $D_j$, $j \in \{1, \cdots b\}$.
For example, in \cref{fig2}, the blue box of $D_1$ represents the bucket of $\boldsymbol{h}_{1,\boldsymbol{y}_m}$.
Since the hash codes of the training data stored in this bucket are similar to $\boldsymbol{h}_{1,\boldsymbol{y}_m}$, their original features are treated as similar candidates of $\boldsymbol{y}_m$.

For simplicity, suppose that $H_j$ is a stack of $r$ random vectors. Thus, the $i$-th ($i \in \{1, \cdots r\}$) bit of $\boldsymbol{h}_{j,\boldsymbol{y}_m}$ is a dot product of the $i$-th random vector in $H_j$ and $\boldsymbol{y}_m$. Suppose a training vector $\boldsymbol{x}_n$ that makes an angle $\alpha$ with $\boldsymbol{y}_m$:
\begin{equation}
	\alpha(\boldsymbol{y}_m, \boldsymbol{x}_n) = \arccos \frac{\boldsymbol{y}_m \cdot \boldsymbol{x}_n} {|\boldsymbol{y}_m| \times |\boldsymbol{x}_n|} \in [0, \pi] .
	\label{eq1}
\end{equation}

The probability that the $i$-th bit of $bin(\boldsymbol{h}_{j,\boldsymbol{y}_m})$ and $bin(\boldsymbol{h}_{j,\boldsymbol{x}_n})$ has the equal value is:
\begin{equation}
	\mathcal{P} (bin(\boldsymbol{h}_{j,\boldsymbol{y}_m})^{<i>} \equiv bin(\boldsymbol{h}_{j,\boldsymbol{x}_n})^{<i>}) = \frac{\pi - \alpha}{\pi} .
	\label{eq2}
\end{equation}

Therefore, in $D_j$, the probability of $\boldsymbol{x}_n$ being the similar candidate of $\boldsymbol{y}_m$ (\ie the two binary codes being identical) is:
\begin{equation}
	\mathcal{P} (bin(\boldsymbol{h}_{j,\boldsymbol{y}_m}) \equiv bin(\boldsymbol{h}_{j,\boldsymbol{x}_n})) = (\frac{\pi - \alpha}{\pi})^r .
	\label{eq3}
\end{equation}

Considering $b$ hash tables, we take $\boldsymbol{x}_n$ as a similar candidate of $\boldsymbol{y}_m$ if there is at least a table that meets the requirement of $bin(\boldsymbol{h}_{j,\boldsymbol{y}_m}) \equiv bin(\boldsymbol{h}_{j,\boldsymbol{x}_n})$, $j \in \{1, \cdots b\}$.
Its probability $\mathcal{P}_{r,b}$ is:
\begin{equation}
	\mathcal{P}_{r,b}(\alpha) = 1 - (1 - (\frac{\pi-\alpha}{\pi})^r)^b .
	\label{eq4}
\end{equation}

\begin{figure}[t]
	\centering
	\includegraphics{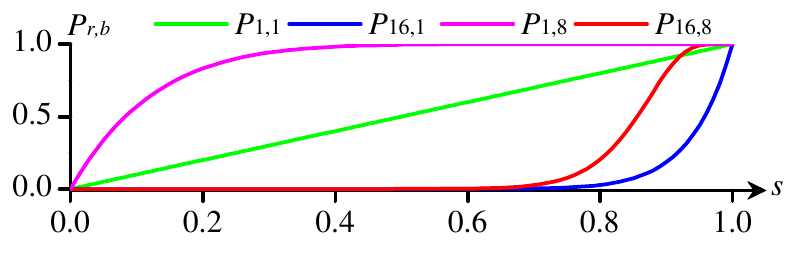}
	\caption{The $P_{r,b}$-$s$ curve. The $(r,b)$ values of green, blue, magenta and red curves are $(1,1)$, $(16,1)$, $(1,8)$ and $(16,8)$ respectively. Best viewed in color.
	}
	\label{fig3}
\end{figure}

We use $s = (\pi-\alpha)/\pi$ to denote the similarity of $\boldsymbol{y}_m$ and $\boldsymbol{x}_n$.
The $\mathcal{P}_{r,b}$-$s$ curves of four specific $(r,b)$ values are shown in \cref{fig3}, where $(r,b)=\{(1,1), (16,1), (1,8), (16,8)\}$.
In the curves for $b>1$ and $r>1$, the similarity threshold corresponding to the steepest rise, determines whether $\boldsymbol{y}_m$ and $\boldsymbol{x}_n$ are similar or not. 
An approximation to the threshold \cite{MiningMassive2020leskovec} is:
\begin{equation}
	\hat{s}(r,b) = (\frac{1}{b})^{\frac{1}{r}} .
	\label{eq5}
\end{equation}

Therefore, we can change the similarity threshold $\hat{s}$ by changing the length of hash codes $r$ and the number of hash layers $b$. If the similarity between $\boldsymbol{y}_m$ and $\boldsymbol{x}_n$ is smaller than $\hat{s}$, $\boldsymbol{x}_n$ can hardly become a similar candidate of $\boldsymbol{y}_m$. Otherwise, it can become the similar candidate with a high probability.

In a vanilla locality-sensitive hashing, we can find all the similar candidates of $\boldsymbol{y}_m$ in $D_1, \cdots D_b$ and compute the distances between the original training features and $\boldsymbol{y}_m$. 
However, for further reducing computation cost, we directly compute the average distance between $\boldsymbol{h}_{j,\boldsymbol{y}_m}$ and the hash codes with the same key $bin(\boldsymbol{h}_{j,\boldsymbol{y}_m})$ in $D_j$ for all $j \in \{1, \cdots b\}$.
If $bin(\boldsymbol{h}_{j,\boldsymbol{y}_m})$ is not in $D_j$, the distance is assigned with a large value.
Finally, the minimum distance among all hash tables is taken as the anomaly score for $\boldsymbol{y}_m$.

\section{Experiments}

\subsection{Datasets and Setups}

\subsubsection{Datasets}
We evaluate our method on three benchmark datasets and compare the performance with the state of the art.

\textbf{Avenue}. The CUHK Avenue dataset \cite{AbnormalEvent2013lu} consists of 16 training and 21 testing videos of a single scene. There are 47 abnormal events in the testing videos, such as loitering, wrong direction, and throwing stuff. The resolution of each video is 360 $\times$ 640 pixels.

\textbf{ShanghaiTech}. The ShanghaiTech (\textit{abbr.} ST) dataset \cite{RevisitSparse2017luo} contains 330 training and 107 testing videos of 13 scenes. Examples of anomalous events are riding bikes, fighting and vehicles. The resolution of each video is 480 $\times$ 856 pixels. It is the largest dataset among existing benchmarks for video anomaly detection.

\textbf{Corridor}. The IITB Corridor dataset \cite{MultitimescaleTrajectory2020rodrigues} contains 208 training and 150 testing videos of a single scene. Anomalous events include protest, hiding and playing with ball. The resolution of each video is 1920 $\times$ 1080 pixels. It is a new and large benchmark for video anomaly detection.

\subsubsection{Evaluation Metrics}
The area under curve (AUC) is the most commonly used evaluation metric for video anomaly detection in frame-level. It is computed by the area under the receiver operating characteristic curve with varying thresholds for anomaly scores. A higher value indicates a better performance.

Following the protocol of \cite{BackgroundAgnosticFramework2021georgescua}, we evaluate the performance with two clearly defined frame-level AUCs as metrics: 1) macro-averaged AUC (\textit{abbr.} macro-AUC), which first computes the frame-level AUC for each video, then averages the resulting AUCs of videos, and 2) micro-averaged AUC (\textit{abbr.} micro-AUC), which first concatenates frame-level scores of all videos and then computes the AUC.
For a fair comparison, the results of different methods are reported under these two evaluation metrics and are compared following the same protocol.
In our ablation studies, we use the macro-AUC for evaluation.

\subsubsection{Implementation Details}
We use the SlowFast network \cite{SlowFastNetworks2019feichtenhofer} pre-trained on Kinetics-400 dataset \cite{QuoVadis2017carreira} to extract features \cite{PySlowFast2020fan}.
Training snippets are sampled from random starting positions in videos.
Every snippet consists of $T=32$ frames with the sampling rate $s=1$. The sampling offset $\varDelta t$ is a random integer in the range of $-150$ and $150$. All the frames are scaled to $256 \times 256$ pixels before fed into the network.
We use $b=8$ hash layers, each of which outputs an $r=32$-bit hash code.
On ShanghaiTech and Corridor datasets, the length of the queue for each dataset is $l=8192$, and the batch size is $256$. On Avenue dataset, $l$ is set to $2048$ and the batch size is $32$.
The models are trained with a learning rate of $0.001$ for $60$, $60$ and $10$ iterations on Avenue, ShanghaiTech and Corridor datasets, respectively.
The momentum coefficient $m$ for $\theta_k$ is set to $0.999$, and the temperature $\tau$ in the \textit{InfoNCE} loss is $0.2$.
Our experiments are conducted with four Nvidia RTX-2080Ti GPUs using PyTorch \cite{AutomaticDifferentiation2017paszke}.
We follow the previous works \cite{ObjectCentricAutoEncoders2019ionescu, AnomalyDetection2021georgescu} to apply a Gaussian filter to temporally smooth the anomaly scores.

\subsection{Comparison with the state of the art}

\begin{table}[t]
	\caption{Quantitative comparison with state-of-the-art methods for anomaly detection.
		Numbers in bold: the best performance; * (also in gray text): object-level methods; \textdagger: our re-implementations; \textdaggerdbl: results from other papers (no official implementations).}
	\centering
	\setlength\tabcolsep{5pt}
	\begin{tabular}{@{}lclccc@{}}
		\toprule	
		\multicolumn{1}{c}{}                                            & Year        & Method       									& Avenue 	& ST   		& Corridor \\ \midrule
		\multirow{16}{*}{\begin{tabular}[c]{@{}l@{}}\rotatebox{90}{Micro-AUC} \end{tabular}}
		& 2018 	 																	  & FFP \cite{FutureFrame2018liu}           		& 85.1		& 72.8		& 64.7\textsuperscript{\textdaggerdbl}       \\  \cmidrule(l){2-6}
		& \multirow{4}{*} {\begin{tabular}[c]{@{}c@{}}2019\end{tabular}}	  		  & AMC \cite{AnomalyDetection2019nguyen}          	& 86.9   	& -    		& -        \\
		&																			  & LSA \cite{LatentSpace2019abati}					& -			& 72.5		& -		   \\
		&                                                                             & MPED-RNN \cite{LearningRegularity2019morais}	& -      	& 73.4 		& 64.3\textsuperscript{\textdaggerdbl}       \\
		&                                                                             & MemAE \cite{MemorizingNormality2019gong}        & 83.3   	& 71.2 		& -        \\ \cmidrule(l){2-6} 
		& \multirow{6}{*}{2020}                                                       & PoseCVAE \cite{PoseCVAEAnomalous2021jain}     	& 82.1   	& 74.9 		& 67.3     \\
		&                                                                             & MNAD \cite{LearningMemoryGuided2020park}		& 88.5		& 70.5		& -        \\
		&                                                                             & MNAD \cite{LearningMemoryGuided2020park} \textsuperscript{\textdagger} 		& 86.9   & 67.5 & 68.1     \\
		&                                                                             & GEPC \cite{GraphEmbedded2020markovitz}        	& -      	& 76.1 		& -        \\
		&                                                                             & MTP \cite{MultitimescaleTrajectory2020rodrigues}& 82.9   	& 76.0 		& 67.1     \\
		&                                                                             & CDDA \cite{ClusteringDriven2020chang}        	& 86.0   	& 73.3 		& -        \\ \cmidrule(l){2-6} 
		& \multirow{4}{*}{2021}                                                       & AMMC-Net \cite{AppearanceMotionMemory2021cai}   & 86.6   	& 73.7 		& -        \\
		&                                                                             & \textcolor[rgb]{0.5,0.5,0.5}{HF\textsuperscript{2}-VAD} \cite{HybridVideo2021liu} \textsuperscript{*} & \textcolor[rgb]{0.5,0.5,0.5}{91.1}   & \textcolor[rgb]{0.5,0.5,0.5}{76.2} & \textcolor[rgb]{0.5,0.5,0.5}{-}        \\
		&                                                                             & MPN \cite{LearningNormal2021lv}         		& \textbf{89.5} & 73.8	& -        \\
		&                                                                             & MPN \cite{LearningNormal2021lv} \textsuperscript{\textdagger} & 83.9 & 73.0 & 69.2     \\ \cmidrule(l){2-6} 
		& \multirow{2}{*}{Ours}                                                       & LSH   											& 86.5   	& 72.1 		& 73.4     \\
		&                                                                             & LLSH  											& 87.4   	& \textbf{77.6} & \textbf{73.5}     \\ \midrule
		\multirow{6}{*}{\begin{tabular}[c]{@{}l@{}}\rotatebox{90}{Macro-AUC}\end{tabular}}  
		& 2019                                                                        & \textcolor[rgb]{0.5,0.5,0.5}{OADA} \cite{ObjectCentricAutoEncoders2019ionescu} \textsuperscript{*}   & \textcolor[rgb]{0.5,0.5,0.5}{90.4}   & \textcolor[rgb]{0.5,0.5,0.5}{84.9} & \textcolor[rgb]{0.5,0.5,0.5}{-}        \\ \cmidrule(l){2-6} 
		& 2020                                                                        & MNAD \cite{LearningMemoryGuided2020park} \textsuperscript{\textdagger} 		& 80.1   & 81.7 & 68.4     \\ \cmidrule(l){2-6} 
		& \multirow{2}{*}{2021}                                                       & MPN \cite{LearningNormal2021lv} \textsuperscript{\textdagger}  & 81.5   & 80.2 & 66.1     \\
		&                                                                             & MTL \cite{AnomalyDetection2021georgescu}        & 86.9   	& 83.5 		& -        \\ \cmidrule(l){2-6} 
		& \multirow{2}{*}{Ours}                                                       & LSH          									& 86.3   	& 80.5 		& 73.8     \\
		&                                                                             & LLSH         									& \textbf{88.6}   & \textbf{85.9} & \textbf{74.2}     \\ \bottomrule
	\end{tabular}
	\label{table1}
\end{table}
We compare our methods using both micro-AUC and macro-AUC evaluation protocols with the state of the art for video anomaly detection on Avenue \cite{AbnormalEvent2013lu}, ShanghaiTech \cite{RevisitSparse2017luo} and Corridor \cite{MultitimescaleTrajectory2020rodrigues} datasets, whose results can be seen in \cref{table1}.
We re-implement MNAD \cite{LearningMemoryGuided2020park} and MPN \cite{LearningNormal2021lv} methods on the three datasets with the help of their official codes, which are marked with \textdagger. 
Rodrigues \etal \cite{MultitimescaleTrajectory2020rodrigues} have implemented FFP \cite{FutureFrame2018liu} and MPED-RNN \cite{LearningRegularity2019morais} on Corridor dataset, so we just report their results, which are marked with \textdaggerdbl.
HF\textsuperscript{2}-VAD \cite{HybridVideo2021liu} and OADA \cite{ObjectCentricAutoEncoders2019ionescu} are two object-level methods, which are marked with * and in gray text. It should be noted that a comparison between object-level methods and frame-level methods are not fair, for the reason that some anomalies (\eg bikes, skateboards and vehicles) can be recognized by object detectors.
Therefore, we only report the frame-level AUC results of MTL \cite{AnomalyDetection2021georgescu}, and do not take HF\textsuperscript{2}-VAD and OADA into account for comparisons of AUC.
Our LSH does not have the training phase. Its parameters are just randomly initialized. LLSH is trained as we introduced in our Method Section.

\begin{figure}[t]
	\centering
	\includegraphics{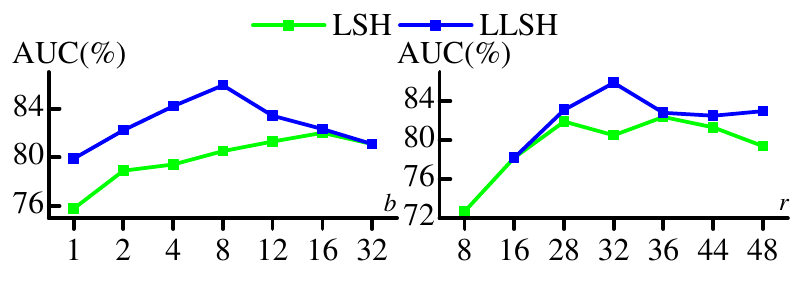}
	\caption{Results of different numbers of hash layers $b$ and different lengths of hash codes $r$ on ST dataset in macro-AUC metric.
	}
	\label{figbr}
\end{figure}

From the table, we observe three things:

(1) A simple LSH method can achieve good performance in VAD. We can see that our LSH method can surpass several recent methods, such as CDDA \cite{ClusteringDriven2020chang}, MTP \cite{MultitimescaleTrajectory2020rodrigues}, PoseCVAE \cite{PoseCVAEAnomalous2021jain}, MemAE \cite{MemorizingNormality2019gong} and FFP \cite{FutureFrame2018liu}.

(2) The performance of LLSH is significantly better than LSH, which demonstrates the effectiveness of making the parameters of LSH learnable. The intra-class variations of ShanghaiTech are much larger than Avenue and Corridor datasets since it has 13 different scenes. However, compared with LSH, the improvement of LLSH on that dataset is the largest, indicating that the process of learning improves the adaptability of LSH.
Due to the limitation of the input size of the pre-trained CNN, large-scale reduction of the high resolution frames is probably the reason of small improvement on Corridor dataset.

(3) Our proposed LLSH method achieves the best results on the two large-scale datasets, \ie, ShanghaiTech and Corridor in both micro-AUC and macro-AUC evaluation protocols.
Its performance on Avenue dataset is also the best in macro-AUC metric, and the third best in micro-AUC metric among frame-level methods.
On ShanghaiTech dataset, LLSH outperforms recent memory-augmented prediction methods \cite{MemorizingNormality2019gong, LearningMemoryGuided2020park, AppearanceMotionMemory2021cai, LearningNormal2021lv} by $4\% \sim 7\%$. As a frame-level method, it even outperforms object-level methods \cite{ObjectCentricAutoEncoders2019ionescu, HybridVideo2021liu} in both metrics consistently.
The outstanding performance demonstrates the superiority of our LLSH.

\subsection{Ablation Study}

To study the effects of the number of hash layers $b$ and the length of hash codes $r$, we perform ablation experiments for the two hyper-parameters on ShanghaiTech dataset. In the study of $b$, we empirically set $r$ to 32, and in the study of $r$, $b$ is empirically set to 8.
Their results in macro-AUC metric are shown in \cref{figbr}.

With the increment of $b$, AUC increases first but then decreases, whose trend resembles that of $r$.
\cref{eq5} explains the reason of this phenomenon. As $r$ is fixed, increment of $b$ makes the similarity threshold decline. As a result, dissimilar hash codes are put in the same bucket, causing high false positive rate. As $b$ is fixed, increment of $r$ makes the similarity threshold rise, causing low true positive rate since similar hash codes cannot be put in the same bucket. A way to improve accuracy is increasing $b$ and $r$ simultaneously. However, the cost of computation also increases linearly as one of $b$ and $r$ increases. In our experiments, $b=8$ and $r=32$ is a proper setting for our model and does not bring much computation cost.

\subsection{Visualization}

\begin{figure}[t]
	\centering
	\includegraphics{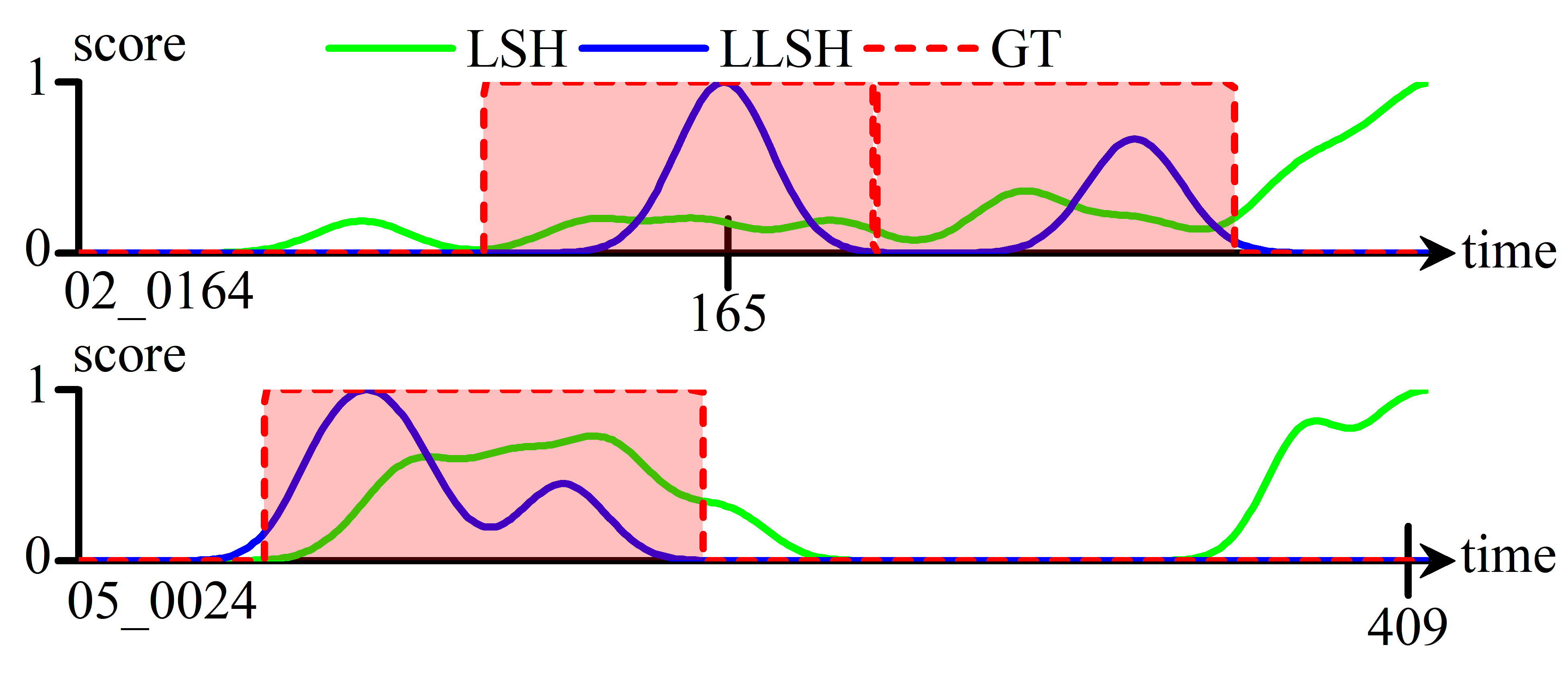}
	\caption{Anomaly score curves of LSH and LLSH on "02\_0164" and "05\_0024" videos from ST dataset. GT stands for groundtruth. A higher score means a higher probability of anomaly.
	}
	\label{figscr}
\end{figure}

\begin{figure}[t]
	\centering
	\includegraphics[width=\linewidth]{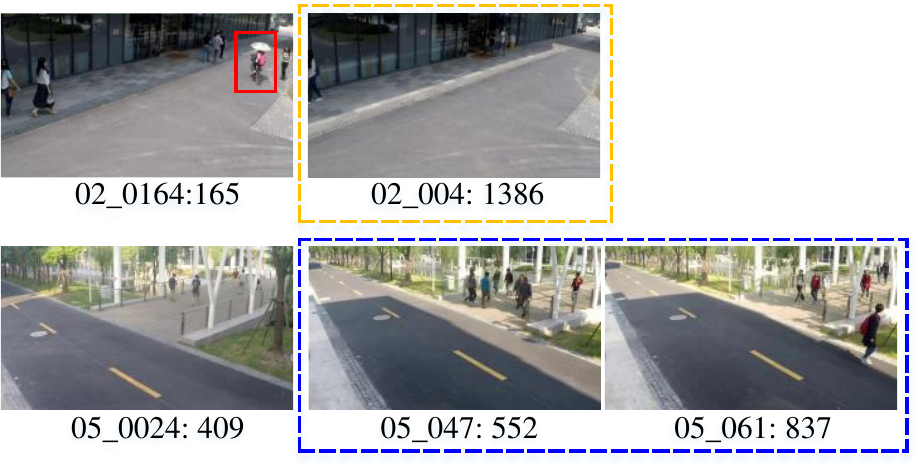}
	\caption{Query snippets (in the first column) and their similar candidate snippets found by LSH (in the orange dashed box) and LLSH (in the blue dashed box). The region in red box represents anomalous region.
	}
	\label{figsamp}
\end{figure}


\begin{table*}[t]
	\centering
	\caption{Comparisons of different methods in performance, computation cost and model size on ShanghaiTech dataset. $d$=9216: dimension of feature; $N$=792855, $M$=112422: numbers of training and testing features; $t$: the maximum iterations; $n$: number of training hash codes to calculate distance with testing hash codes in query stage; $m$: number of training hash codes to calculate mean vectors in index stage.
	}
	\label{tabeff}
	\resizebox{0.98\textwidth}{!}{%
		\begin{tabular}{l|c|cc|cc|c}
			\hline
			& KNN                     & \multicolumn{2}{c|}{K-means}                 & \multicolumn{1}{c|}{LSH}                                                                              & LLSH                                                                              & light-LLSH                                                                                \\ \hline
			\multirow{2}{*}{Parameters}                                                          & \multirow{2}{*}{$K$=1024} & \multicolumn{2}{c|}{$t$=300}                   & \multicolumn{1}{c|}{\multirow{2}{*}{\begin{tabular}[c]{@{}c@{}}$b$=8, $r$=32,\\ $n$=38150830\end{tabular}}} & \multirow{2}{*}{\begin{tabular}[c]{@{}c@{}}$b$=8, $r$=32,\\ $n$=209872075\end{tabular}} & \multirow{2}{*}{\begin{tabular}[c]{@{}c@{}}$b$=8, $r$=32, $n$=490433,\\ $m$=5650569\end{tabular}} \\ \cline{3-4}
			&                         & \multicolumn{1}{c|}{$K$=1024}      & $K$=32      & \multicolumn{1}{c|}{}                                                                                 &                                                                                   &                                                                                           \\ \hline
			macro-AUC                                                                           & 85.6\%                    & \multicolumn{1}{c|}{84.1\%}        & 83.3\%      & \multicolumn{1}{c|}{80.5\%}                                                                             & 85.9\%                                                                              & 85.7\%                                                                                      \\ \hline
			\multirow{2}{*}{\begin{tabular}[c]{@{}l@{}}Number of\\ multiplications\end{tabular}} & $dNM$                     & \multicolumn{2}{c|}{$dKNt+dKM$}                & \multicolumn{2}{c|}{$drb(M+N)+rn$}                                                                                                                                                          & $drb(M+N)+2rm+rn$                                                                           \\ \cline{2-7} 
			& 821.5 Tera              & \multicolumn{1}{c|}{2245.8 Tera} & 70.2 Tera & \multicolumn{1}{c|}{2.1 Tera}                                                                         & 5.5 Giga more than LSH                                                            & 0.8 Giga less than LSH                                                                    \\ \hline
			Disk   usage                                                                         & 28 GB                       & \multicolumn{1}{c|}{111   MB}    & 6.4   MB  & \multicolumn{1}{c|}{1190   MB}                                                                        & 993   MB                                                                          & 316   MB                                                                                  \\ \hline
		\end{tabular}%
	}
\end{table*}


We visualize two samples from ShanghaiTech dataset to qualitatively analyze how LLSH improves the accuracy for video anomaly detection compared with LSH.
The anomaly score curves of LSH and LLSH methods on "02\_0164" and "05\_0024" videos from ShanghaiTech dataset are shown in \cref{figscr}. After the process of learning hashing, the AUC on "02\_0164" rises from 64.1\% (LSH) to 93.7\% (LLSH), and rises from 80.2\% (LSH) to 99.5\% (LLSH) on "05\_0024".
From the curves we can see that LLSH can increase anomaly scores for anomalous frames and decrease anomaly scores for normal frames, hence improving the accuracy for video anomaly detection.

We take two query snippets which are centered at the 165-th frame of "02\_0164" and the 409-th frame of "05\_0024" as examples, to visualize how LLSH modifies anomaly scores.
For simplicity, we use the center frame to represent a snippet.
In the first row of \cref{figsamp}, the snippet in orange dashed box is the only similar candidate found by LSH. The cosine similarity between the features of it and the corresponding query snippet is 0.8786, generating an anomaly score of 0.1756. However, the 165-th frame of "02\_0164" is anomalous, which expects a high score. The LLSH method does not find any similar candidates in all the hash tables, thus generating a high score of 0.9965.
The 409-th frame of "05\_0024" is normal and expects a low score. However, LSH finds no similar candidate and hence generates a high score of 0.9719. In contrast, LLSH finds a total of 9 similar candidates from 3 videos in the nearest bucket. Two of the similar snippets are show in the second row of \cref{figsamp} in blue dashed box. The average cosine similarity between this query snippet and those 9 similar candidates is 0.8880, according to which LLSH generates a low anomaly score of 0.0001.
From the above two examples, we can see that compared with LSH, LLSH can eliminate wrong similar candidates for anomalous queries and find correct similar candidates for normal queries. Therefore, LLSH improves the accuracy for video anomaly detection.

\subsection{Efficiency}
We conduct a series of experiments using KNN and K-means \cite{ScikitlearnMachine2011pedregosa} methods for video anomaly detection, and select the models with high AUC to compare with our proposed methods in performance (macro-AUC), computation cost in the testing phase (number of multiplications) and model size (disk usage).
For a testing feature, KNN uses the average distance of the $K$ nearest training features as the anomaly score. K-means adopts the nearest distance between the testing feature and the $K$ cluster centers as the anomaly score.
Moreover, we propose a light version of LLSH (light-LLSH), which calculates the mean vector of each bucket in the index stage, \ie, every bucket only contains a mean vector.
Comparisons on ShanghaiTech dataset are shown in \cref{tabeff}.
It can be seen that:
(1) Simple KNN and K-means methods can achieve good performance in anomaly detection.
(2) Our LSH, LLSH and light-LLSH require much less computation cost than KNN and K-means, demonstrating the efficiency of our methods.
Specifically, the computation cost of LLSH is only $0.26\%$ of KNN and $3.00\%$ of K-means ($K$=32).
(3) The proposed light-LLSH has a slightly worse performance but much less computation cost and model size than LLSH. Although K-means has the smallest model size, its performance is significantly worse than light-LLSH and causes large computation cost.
Overall, our light-LLSH makes the best trade-off among performance, computation cost and model size.

\subsection{Adaptability}

\begin{figure}[t]
	\centering
	\includegraphics[]{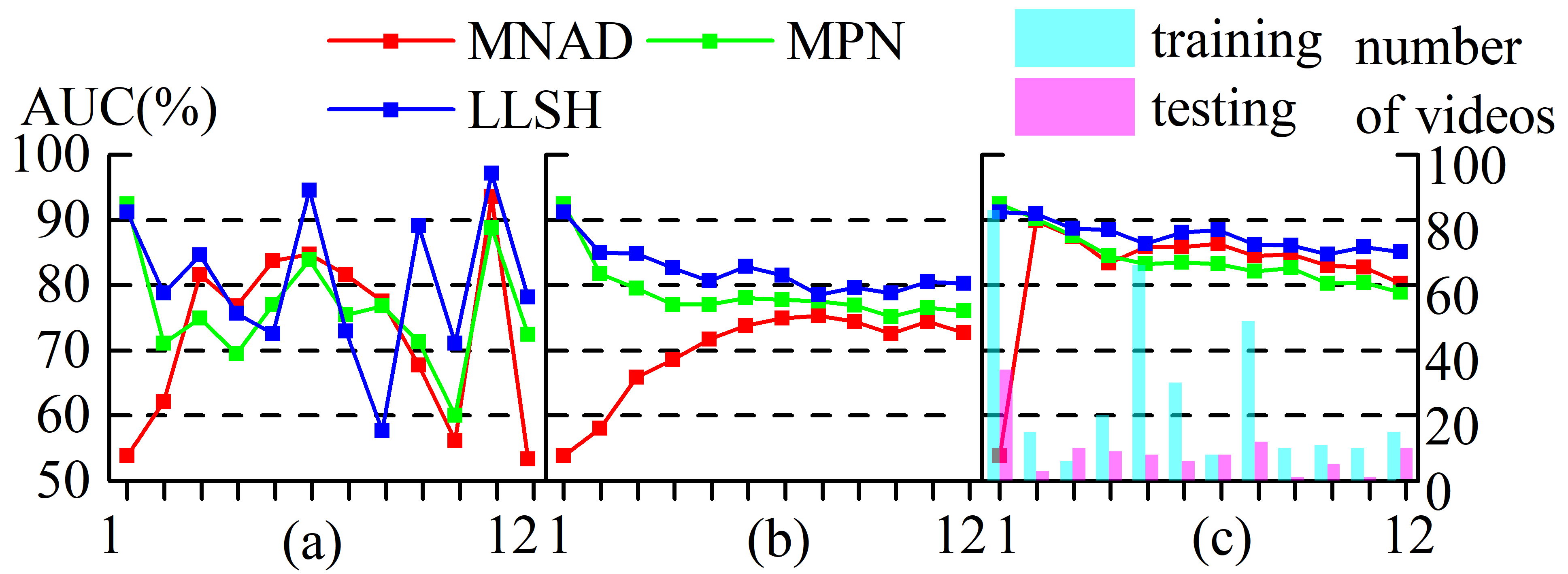}
	\caption{Results (macro-AUC) of MNAD, MPN and LLSH under three circumstances, and video numbers in each scene.
	}
	\label{figadp}
\end{figure}

ShangheiTech is the most challenging dataset since it has large intra-class variations and imbalanced distribution. Therefore, we use it to study the adaptability of different methods.
We compare MNAD \cite{LearningMemoryGuided2020park}, MPN \cite{LearningNormal2021lv} and our LLSH under three circumstances:
(a) training and testing models in each single scene,
(b) computing average result of the first $N$ scenes obtained in (a),
and (c) training and testing models in the first $N$ scenes.
Although there are 13 scenes in the training data, there are 12 scenes in the testing data. Hence we use the 12 common scenes in the experiments, \ie, $N \in \{1, 2 \cdots, 12\}$.
\cref{figadp} shows the results under these three circumstances and numbers of training/testing videos in each scene.
From the results in (a) and (b) we can see that LLSH outperforms MNAD and MPN in most scenes consistently.
In the 8-th scene, jogging in the crowd causes occlusion and breaks the continuity of motion, thus our freezed pre-trained CNN cannot extract discriminative representations for some snippets. For this reason, LLSH gets a lower result in this scene.
The results of (c) demonstrates the super adaptability of LLSH since it can always achieve the best performance with the increment of new scenes, even if video numbers change largely in different scenes.
Besides, comparing the results of (b) and (c), we find that all the three methods can benefit from training in multiple scenes. That is to say, normal information between different scenes can conduce to complement each other.

\subsection{Scalability}
\begin{figure}[t]
	\centering
	\includegraphics[]{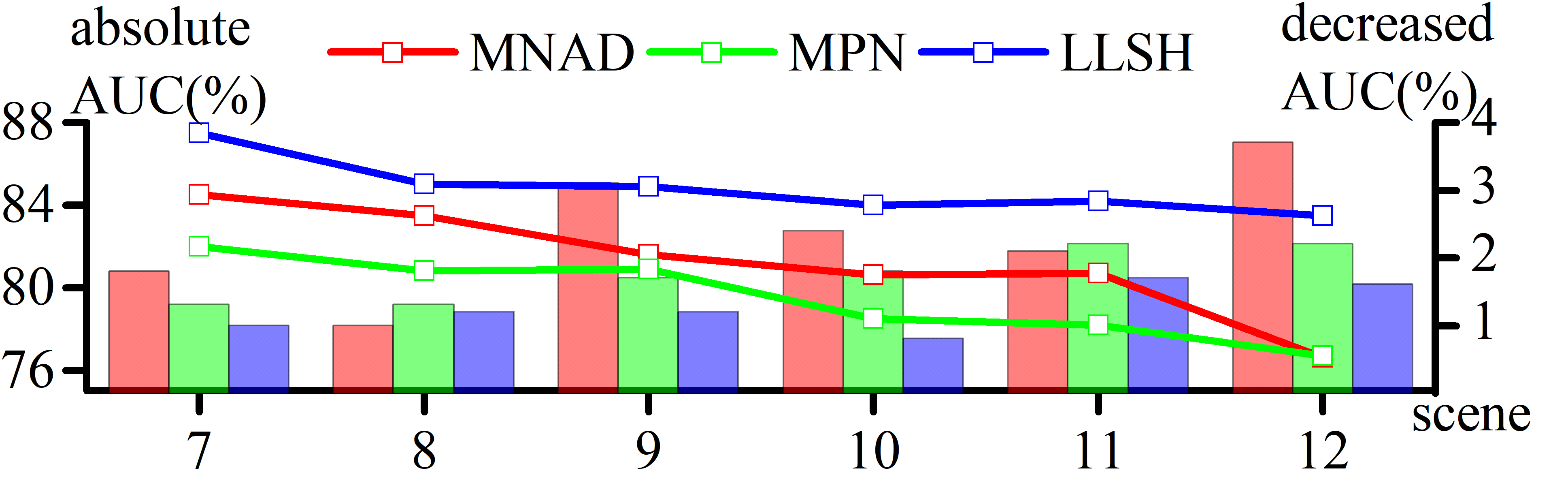}
	\caption{Results (macro-AUC) of MNAD, MPN and LLSH for newly added scenes on ST dataset. The lines/bars are absolute/decreased AUCs on the first $7,8, \cdots, 12$ scenes, where a higher value means better/worse respectively.
	}
	\label{figsca}
\end{figure}

A model with strong scalability has more practical application value.
Therefore, we compare the scalability of MNAD \cite{LearningMemoryGuided2020park}, MPN \cite{LearningNormal2021lv} and the proposed LLSH for newly added scenes.
For each model, we train it on the first 6 scenes of ShanghaiTech dataset, and directly test it on the first $7, 8, \cdots, 12$ scenes. 
The results are shown in \cref{figsca}, where absolute AUCs are plotted in lines, and decreased AUCs compared with training with all the scenes together are plotted in bars.
LLSH consistently achieves the best performance in absolute AUC for different numbers of newly added scenes. In addition, it nearly always has the least decline of performance, which indicates its superior of scalability.
Our proposed LLSH has the advantages of both learnable models and non-learnable algorithms, making it adaptive and also retaining its scalability.

\section{Conclusion}
In this paper, we have proposed a novel learnable locality-sensitive hashing (LLSH), which can take full advantage of the knowledge of normal data efficiently and flexibly, to distinguish anomalies from normal events.
It implements LSH as a parametric network and trains end-to-end in a contrastive learning framework.
Extensive experiments have shown that LLSH is superior in efficiency, adaptability and scalability, achieving new state of the art on VAD benchmarks. 


{\small
\bibliographystyle{ieeefullname}
\bibliography{VADrefs}
}

\end{document}